\documentclass[runningheads,a4paper]{llncs}
\usepackage[english]{babel}
\usepackage{proof}
\usepackage{hyperref}
\usepackage{stmaryrd}
\usepackage[figuresleft]{rotating}

\newcommand{\negl}[1]{{^0#1}}
\newcommand{\negr}[1]{{#1^0}}
\newcommand{\conegl}[1]{{^1#1}}
\newcommand{\conegr}[1]{{#1^1}}

\newcommand{\termvara}{M}
\newcommand{\termvarb}{N}

\newcommand{\nvara}{\varphi}
\newcommand{\nvarb}{\psi}
\newcommand{\nvarc}{\rho}

\newcommand{\vara}{x}
\newcommand{\varb}{y}
\newcommand{\varc}{z}

\newcommand{\covara}{\varepsilon}
\newcommand{\covarb}{\kappa}
\newcommand{\covarc}{\nu}

\newcommand{\fusionpair}[2]{\langle #1\lgtimes #2\rangle}
\newcommand{\casefusion}[4]{\textbf{case} \ #1 \ \textbf{of} \ #2\lgtimes #3.#4}

\newcommand{\lsem}{\llbracket}
\newcommand{\rsem}{\rrbracket}
\newcommand{\sem}[1]{\lsem #1\rsem}
\newcommand{\floor}[1]{\lfloor #1\rfloor}
\newcommand{\ceil}[1]{\lceil #1\rceil}

\newcommand{\result}{\bot}

\newcommand{\leftspoon}{\mathbin{\circ-}}
\newcommand{\rightspoon}{\mathbin{-\circ}}
\newcommand{\leftfilledspoon}{\mathbin{\bullet-}}
\newcommand{\rightfilledspoon}{\mathbin{-\bullet}}

\newcommand{\np}{\textit{np}}

\newcommand{\bs}{\backslash}
\newcommand{\os}{\varoslash}
\newcommand{\obs}{\varobslash}
\newcommand{\lgtimes}{\varotimes}
\newcommand{\lgplus}{\varoplus}

\newcommand{\seq}[3]{#1,#2\vdash #3}
\newcommand{\seqs}[3]{#1\vdash #2:#3}
\newcommand{\seqlp}[3]{#1\vdash #2:#3}
\newcommand{\seqlpd}[3]{#1\vdash_{\textbf{LP}} #2:#3}

\newcommand{\stimes}{\bullet}
\newcommand{\splus}{\circ}
\newcommand{\ssl}{\rightspoon}
\newcommand{\sbs}{\leftspoon}
\newcommand{\sos}{\leftfilledspoon}
\newcommand{\sobs}{\rightfilledspoon}
\newcommand{\slgc}[1]{{#1^{\rightspoon}}}
\newcommand{\srgc}[1]{{^{\leftspoon}#1}}
\newcommand{\sldgc}[1]{{^{\leftfilledspoon}#1}}
\newcommand{\srdgc}[1]{#1^{\rightfilledspoon}}

\title{Polarized Montagovian semantics for the Lambek-Grishin calculus}
\author{Arno Bastenhof}
\institute{Utrecht University}

\begin{document}
\bibliographystyle{plain}
\maketitle
\begin{abstract}
Grishin (\cite{grishin}) proposed enriching the Lambek calculus with multiplicative disjunction (par) and coresiduals. Applications to linguistics were discussed by Moortgat (\cite{moortgat09}), who spoke of the Lambek-Grishin calculus (\textbf{LG}). In this paper, we adapt Girard's polarity-sensitive double negation embedding for classical logic (\cite{girard91}) to extract a compositional Montagovian semantics from a display calculus for focused proof search (\cite{andreoli92}) in \textbf{LG}. We seize the opportunity to illustrate our approach alongside an analysis of extraction, providing linguistic motivation for linear distributivity of tensor over par (\cite{cockettseely}), thus answering a question of \cite{kurtoninamoortgat}. We conclude by comparing our proposal to that of \cite{bernardimoortgat}, where alternative semantic interpretations of \textbf{LG} are considered on the basis of call-by-name and call-by-value evaluation strategies.
\end{abstract}
Inspired by Lambek's syntactic calculus, Categorial type logics (\cite{moortgat97}) aim at a proof-theoretic explanation of natural language syntax: syntactic categories and grammaticality are identified with formulas and provability. Typically, they show an intuitionistic bias towards asymmetric consequence, relating a structured configuration of hypotheses (a constituent) to a single conclusion (its category). The \textit{Lambek-Grishin calculus} (\textbf{LG}, \cite{moortgat09}) breaks with this tradition by restoring symmetry, rendering available (possibly) multiple conclusions. $\S$1 briefly recapitulates material on \textbf{LG} from \cite{moortgat09} and \cite{moortgat10}. 

In this article, we couple \textbf{LG} with a Montagovian semantics.\footnote{Understanding \textit{Montagovian semantics} in a broad sense, we take as its keywords \textit{model-theoretic} and \textit{compositional}. Our emphasis in this article lies on the latter.} Presented in $\S$2, its main ingredients are focused proof search \cite{andreoli92} and a double negation translation along the lines of \cite{girard91} and \cite{lafontstreicherreus}, employing \textit{polarities} to keep the number of negations low. In $\S$3, we illustrate our semantics alongside an analysis of extraction inspired by linear distributivity principles (\cite{cockettseely}). Finally, $\S$4 compares our approach to the competing proposal of \cite{bernardimoortgat}.

\section{The Lambek-Grishin calculus}
Lambek's (non-associative) syntactic calculus (\textbf{(N)L}, \cite{lambek58}, \cite{lambek61}) combines linguistic inquiry with the mathematical rigour of proof theory, identifying syntactic categories and derivations by formulas and proofs respectively. On the logical side, \textbf{(N)L} has been identified as (non-associative, )non-commutative multiplicative intuitionistic linear logic, its formulas generated as follows:
\begin{center}
\begin{tabular}{cclcr}
$A..E$ & $::=$ & $p$ & \ \ \ \ \ & Atoms/propositional variables \\
 & $|$ & $(A\lgtimes B)$ & & Multiplicative conjunction/tensor \\
 & $|$ & $(B\bs A)$ $|$ $(A/B)$ & & Left and right implication/division
\end{tabular}
\end{center}
Among \textbf{NL}'s recent offspring we find the Lambek-Grishin calculus (\textbf{LG}) of \cite{moortgat09}, inspired by Grishin's (\cite{grishin}) extension of Lambek's vocabulary with a multiplicative disjunction (\textit{par}) and \textit{coimplications/subtractions}. Combined with the recent addition of (co)negations proposed in \cite{moortgat10}, the definition of formulas now reads
\begin{center}
\begin{tabular}{cclcr}
$A..E$ & $::=$ & $p$ & \ \ \ \ \ & Atoms \\
 & $|$ & $(A\lgtimes B)$ $|$ $(A\lgplus B)$ & & Tensor vs. par \\
 & $|$ & $(A/B)$ $|$ $(B\obs A)$ & & Right division vs. left subtraction \\
 & $|$ & $(B\bs A)$ $|$ $(A\os B)$ & & Left division vs. right subtraction \\
 & $|$ & $\negl{B}$ $|$ $\conegr{B}$ & & Left negation vs. right conegation \\
 & $|$ & $\negr{B}$ $|$ $\conegl{B}$ & & Right negation vs. left conegation
\end{tabular}
\end{center}
Derivability ($\leq$) satisfies the obvious preorder laws:
\begin{center}
\begin{tabular}{ccc}
$\begin{array}{c}\infer[\textit{Refl}]{A\leq A}{}\end{array}$ & \ \ \ &
$\begin{array}{c}\infer[\textit{Trans}]{A\leq C}{A\leq B & B\leq C}\end{array}$
\end{tabular}
\end{center}
Logical constants group into families with independent algebraic interest. The connectives $\{\lgtimes,/,\bs\}$ constitute a \textit{residuated family} with \textit{parent} $\lgtimes$, while $\{\lgplus,\obs,\os\}$ embodies the dual concept of a coresiduated family. Finally, $\{\negl{\cdot},\negr{\cdot}\}$ and $\{\conegr{\cdot},\conegl{\cdot}\}$ represent Galois-connected and dually Galois-connected pairs respectively.\footnote{Throughout this text, a double line indicates derivability to go in both ways. Similarly, in $(m)$ below, the same premises are to be understood as deriving each of the inequalities listed under the horizontal line.}
\begin{center}
\begin{tabular}{ccccccc}
$\begin{array}{c}\infer=[r]{A\leq C/B}{\infer=[r]{A\lgtimes B\leq C}{B\leq A\bs C}}\end{array}$ & \ \ \ \ &
$\begin{array}{c}\infer=[cr]{A\obs C\leq B}{\infer=[cr]{C\leq A\lgplus B}{C\os B\leq A}}\end{array}$ & \ \ \ \ \ &
$\begin{array}{c}\infer=[gc]{B\leq\negr{A}}{A\leq\negl{B}}\end{array}$ & \ \ \ \ \ &
$\begin{array}{c}\infer=[dgc]{\conegl{B}\leq A}{\conegr{A}\leq B}\end{array}$
\end{tabular}
\end{center}
Among the derived rules of inference, we find the monotonicity laws:
\begin{center}
\begin{tabular}{ccccccc}
$\begin{array}{c}\infer[m]{\begin{array}{c}A\lgtimes C\leq B\lgtimes D \\ A/D\leq B/C \\ D\bs A\leq C\bs B\end{array}}{A\leq B & C\leq D}\end{array}$ & \ \ \ \ \ &
$\begin{array}{c}\infer[m]{\begin{array}{c}A\lgplus C\leq B\lgplus D \\ A\os D\leq B\os C \\ D\obs A\leq C\obs B\end{array}}{A\leq B & C\leq D}\end{array}$ & \ \ \ \ \ &
$\begin{array}{c}\infer[m]{\begin{array}{c}\negl{B}\leq\negl{A} \\ \negr{B}\leq\negr{A}\end{array}}{A\leq B}\end{array}$ & \ \ \ \ \ &
$\begin{array}{c}\infer[m]{\begin{array}{c}\conegr{B}\leq\conegr{A} \\ \conegl{B}\leq\conegl{A}\end{array}}{A\leq B}\end{array}$
\end{tabular}
\end{center}
\textbf{LG} differs most prominently from \textbf{NL} in the existence of an \textit{order-reversing duality}: an involution $\cdot^{\infty}$ on formulas s.t. $A\leq B$ iff $B^{\infty}\leq A^{\infty}$, realized by\footnote{The present formulation, adopted from \cite{moortgat10}, abbreviates a list of defining equations $(A\lgtimes B)^{\infty}=B^{\infty}\lgplus A^{\infty}$, $(B\lgplus A)^{\infty}=A^{\infty}\lgtimes B^{\infty}$, etc.}
\begin{center}
\begin{tabular}{cccccc}
 \ $p$ \ &
 \ $A\lgtimes B$ \ &
 \ $A/B$ \ &
 \ $B\bs A$ \ &
 \ $\negl{B}$ \ &
 \ $\negr{B}$ \ \\ \hline
 \ $p$ \ &
 \ $B\lgplus A$ \ &
 \ $B\obs A$ \ &
 \ $A\os B$ \ &
 \ $\conegr{B}$ \ &
 \ $\conegl{B}$ \
\end{tabular}$\infty$
\end{center}
A reasonable way of extending \textbf{LG} would be to allow connectives of different families to interact. 
Postulates licensing linear distributivity of $\lgtimes$ over $\lgplus$ (\cite{cockettseely}, \cite{dosenpetric}) come to mind, each being self-dual under $\cdot^{\infty}$ (thus preserving arrow-reversal):
\begin{center}
\begin{tabular}{ccc}
$(A\lgplus B)\lgtimes C\leq A\lgplus(B\lgtimes C)$ & \ \ \ &
$A\lgtimes(B\lgplus C)\leq(A\lgtimes B)\lgplus C$ \\
$(A\lgplus B)\lgtimes C\leq(A\lgtimes C)\lgplus B$ & \ \ \ &
$A\lgtimes(B\lgplus C)\leq B\lgplus(A\lgtimes C)$
\end{tabular}
\end{center}
$\S$3 further explores the relation \textbf{LG}/linguistics, providing as a case analysis for our Montagovian semantics of $\S$2 a sample grammar providing linguistic support for the above linear distributivity principles. As for their proof-theoretic motivation, we note that the following generalizations of Cut become derivable: 
\begin{center}
\begin{tabular}{ccc}
$\infer{A\lgtimes D\leq E\lgplus C}{A\leq E\lgplus B & B\lgtimes D\leq C}$ & \ \ \ \ \ &
$\infer{D\lgtimes A\leq C\lgplus E}{A\leq B\lgplus E & D\lgtimes B\leq C}$ \\ \\
$\infer{A\lgtimes D\leq C\lgplus E}{A\leq B\lgplus E & B\lgtimes D\leq C}$ & \ \ \ \ \ &
$\infer{D\lgtimes A\leq E\lgplus C}{A\leq E\lgplus B & D\lgtimes B\leq C}$
\end{tabular}
\end{center}
For example, suppose $A\leq E\lgplus B$ and $B\lgtimes D\leq C$. From monotonicity and linear distributivity we then deduce
\begin{center}
$A\lgtimes D\leq (E\lgplus B)\lgtimes D\leq E\lgplus(B\lgtimes D)\leq E\lgplus C$
\end{center}

\section{Derivational Montagovian semantics}
We split our semantics into a \textit{derivational} and a \textit{lexical} component. The former is hard-wired into the grammar architecture and tells us how each inference rule builds the denotation of its conclusion (the derived constituent) from those of its premises (the direct subconstituents). The descriptive linguist gives the base of the recursion: the lexical semantics, specifying the denotations of words. Leaving lexical issues aside until $\S$3, we define a $\lambda$-term labeled sequent calculus for simultaneously representing the proofs and the derivational semantics of \textbf{LG}. Summarized in Figure \ref{lgfoc_rules}, its main features (and primary influences) are as follows:
\begin{enumerate}
\item It is, first and foremost, a \textit{display calculus} along the lines of \cite{gore98} and \cite{moortgat09}. In particular, the notion of sequent is generalized so as to accommodate structural counterparts for each of the logical connectives. Display postulates then allow to pick out any hypothesis (conclusion) as the whole of the sequent's antecedent (consequent).
\item Our sequents are labeled by linear $\lambda$-terms for representing compositional meaning construction, adapting the polarity-sensitive double negation translations of \cite{girard91} and \cite{lafontstreicherreus}.
\item We fake a one(/left)-sided sequent presentation to more closely reflect the target calculus of semantic interpretation. To this end, we adapt to our needs de Groote and Lamarche's one(/right)-sided sequents for classical \textbf{NL} (\cite{degrootelamarche}).
\item In contrast with the works cited, our inference rules accommodate focused proof search (\cite{andreoli92}), thus eliminating to a large extent the trivial rule permutations for which sequent calculi are notorious.
\end{enumerate} 
We proceed step by step, starting with a specification of the target language for the double negation translation.
\begin{figure}
\begin{center}
$\infer[Ax]{\seqlp{\tau^{\vara}}{\vara}{\tau}}{}$ \\
\begin{tabular}{ccc} \\
$\infer[\lnot E]{\seqlp{\Gamma,\Delta}{(\termvara \ \termvarb)}{\result}}{
\seqlp{\Gamma}{\termvara}{\lnot \tau} &
\seqlp{\Delta}{\termvarb}{\tau}}$ & \ \ \ \ \ &
$\infer[\lnot I]{\seqlp{\Gamma}{\lambda\vara^{\tau}\termvara}{\lnot \tau}}{
\seqlp{\Gamma,\tau^{\vara}}{\termvara}{\result}}$ \\ \\
$\infer[\lgtimes E]{\seqlp{\Gamma,\Delta}{(\casefusion{\termvarb}{\vara^{\sigma_1}}{\varb^{\sigma_2}}{\termvara})}{\tau}}{
\seqlp{\Delta}{\termvarb}{\sigma_1\lgtimes\sigma_2} &
\seqlp{\Gamma,\sigma_1^{\vara},\sigma_2^{\varb}}{\termvara}{\tau}}$ & \ \ \ \ \ &
$\infer[\lgtimes I]{\seqlp{\Gamma,\Delta}{\fusionpair{\termvara}{\termvarb}}{\tau\lgtimes\sigma}}{
\seqlp{\Gamma}{\termvara}{\tau} &
\seqlp{\Delta}{\termvarb}{\sigma}}$ \\
\end{tabular} \\
\begin{tabular}{rcl} \\
$(\lambda\vara\termvara \ \termvarb)$ & $\rightarrow_{\beta}$ & $\termvara[\termvarb/\vara]$ \\
$\casefusion{\fusionpair{\termvarb_1}{\termvarb_2}}{\vara}{\varb}{\termvara}$ & $\rightarrow_{\beta}$ & $\termvara[\termvarb_1/\vara,\termvarb_2/\varb]$ \\
$((\casefusion{M_1}{x}{y}{M_2}) \ N)$ & $\rightarrow_c$ & $\casefusion{M_1}{x}{y}{(M_2 \ N)}$ \\
$\casefusion{(\casefusion{N_1}{x}{y}{N_2})}{u}{v}{M}$ & $\rightarrow_c$ & $\casefusion{N_1}{x}{y}{\casefusion{N_2}{u}{v}{M}}$
\end{tabular}
\end{center}

\caption{Target language: typing rules and reductions. The $c$-conversions correspond to the obligatory commutative conversions of Prawitz (\cite{prawitz_nd}).}
\label{target_lp}
\end{figure}

\subsubsection{Target language}
Instructions for meaning composition will be phrased in the linear $\lambda$-calculus of Figure \ref{target_lp}, simply referred to as \textbf{LP}. Note that, through the Curry-Howard isomorphism, we may as well speak of a Natural Deduction presentation of multiplicative intuitionistic linear logic. Types $\tau,\sigma$ include multiplicative products $\tau\lgtimes\sigma$ and minimal negations $\lnot\tau$, the latter understood as linear implications $\tau\rightspoon\result$ with result a distinguished atom $\result$:
\begin{center}
$\tau,\sigma$ $::=$ $p$ $|$ $\bot$ $|$ $(\tau\lgtimes\sigma)$ $|$ $\lnot\tau$
\end{center}
We have understood \textbf{LP} to inherit all atoms $p$ of \textbf{LG}. Terms $M$ are typed relative to contexts $\Gamma,\Delta$: multisets $\{\tau_1^{\vara_1},\dots,\tau_n^{\vara_n}\}$ of type assignments $\tau_1,\dots,\tau_n$ to the free variables $\vara_1,\dots,\vara_n$ in $M$. We often omit the braces $\{ \ \}$ and loosely write $\Gamma,\Delta$ for multiset union. Terms in context are represented by \textit{sequents} $\seqlp{\Gamma}{M}{\tau}$, satisfying the linearity constraint that each variable in $\Gamma$ is to occur free in $M$ exactly once. We often write $\seqlpd{\Gamma}{M}{\tau}$ to indicate $\seqlp{\Gamma}{M}{\tau}$ is well-typed.

\subsubsection{From formulas to types: introducing polarities.}
In defining the type $\lsem A\rsem$ associated with a formula $A$, we will parameterize over the following partitioning of non-atomic formulas, speaking of a positive/negative \textit{polarity}:
\begin{center}
\begin{tabular}{lclcr}
Positive(ly polar): & \ \ \ \ \ & $A\lgtimes B,A\os B,B\obs A,\conegr{A},\conegl{A}$ & \ \ \ \ \ & (Metavariables $P,Q$) \\
Negative(ly polar): & \ \ \ \ \ & $A\lgplus B,B\bs A,A/B,\negl{A},\negr{A}$ & & (Metavariables $K,L$)
\end{tabular}
\end{center}
Notice that the dual of a positive formula under $\cdot^{\infty}$ is negative and vice versa, motivating our choice of terminology. In other words, through order reversal, a positive formula on one side of the inequality sign has a negative counterpart behaving alike on the other side. In fact, we shall see that all positive formulas share proof-theoretic behavior, as do all negative formulas. 

We define $\lsem A\rsem$ relative to the polarities $\epsilon(\overrightarrow{B})$ of $A$'s direct subformulae $\overrightarrow{B}$ ($+$ if positive, $-$ if negative). Roughly, a connective expects its polarity to be
\begin{table}
\begin{center}\small
\begin{tabular}{|cc|c|c|c|c|c|c|} \hline
 \ $\epsilon(A)$ \ & \ $\epsilon(B)$ \ & \ \ $\sem{A\lgtimes B}$ \ \ & \ \ $\begin{array}{c}\sem{B\obs A} \\ \sem{A\os B}\end{array}$ \ \ & \ \ $\sem{A\lgplus B}$ \ \ & \ \ $\begin{array}{c}\sem{A/B} \\ \sem{B\bs A}\end{array}$ & $\begin{array}{c}\sem{\negl{B}} \\ \sem{\negr{B}}\end{array}$ & $\begin{array}{c}\sem{\conegr{B}} \\ \sem{\conegl{B}}\end{array}$ \ \ \\ \hline
$-$ & $-$ & $\lnot\sem{A}\lgtimes\lnot\sem{B}$ & $\lnot\sem{A}\lgtimes\sem{B}$ & $\underline{\sem{A}\lgtimes\sem{B}}$ & $\sem{A}\lgtimes\lnot\sem{B}$ & $\lnot\sem{B}$ & $\underline{\sem{B}}$ \\
$-$ & $+$ & $\lnot\sem{A}\lgtimes\sem{B}$ & $\lnot\sem{A}\lgtimes\lnot\sem{B}$ & $\sem{A}\lgtimes\lnot\sem{B}$ & $\underline{\sem{A}\lgtimes\sem{B}}$ & $\underline{\sem{B}}$ & $\lnot\sem{B}$ \\
$+$ & $-$ & $\sem{A}\lgtimes\lnot\sem{B}$ & $\underline{\sem{A}\lgtimes\sem{B}}$ & $\lnot\sem{A}\lgtimes\sem{B}$ & $\lnot\sem{A}\lgtimes\lnot\sem{B}$ & $-$ & $-$ \\
$+$ & $+$ & $\underline{\sem{A}\lgtimes\sem{B}}$ & $\sem{A}\lgtimes\lnot\sem{B}$ & $\lnot\sem{A}\lgtimes\lnot\sem{B}$ & $\lnot\sem{A}\lgtimes\sem{B}$ & $-$ & $-$ \\ \hline
\end{tabular}\normalsize
\end{center}
\caption{Interpreting \textbf{LG}'s formulas by \textbf{LP}'s types.}
\label{polarities}
\end{table}
 preserved by an argument when upward monotone, while reversed when downward monotone. This is the default case, and is underlined for each connective separately in Table \ref{polarities}. Deviations are recorded by marking the offending argument by $\lnot$. In practice, we sometimes loosely refer by $\sem{A}$ to some type isomorphic to it through commutativity and associativity of $\lgtimes$ in \textbf{LP}.

We face a choice in extending the positive/negative distinction to atoms: if assigned positive \textit{bias} (i.e., $\epsilon(p)$ is chosen $+$), $\sem{p}=p$, while $\sem{p}=\lnot p$ with negative bias ($\epsilon(p)=-$). To keep our semantics free from superfluous negations, we go with the former option.

\subsubsection{Antecedent and consequent structures.}
Conforming to the existence of $\cdot^{\infty}$, we consider sequents harboring possible multitudes of both \textit{hypotheses} (or \textit{inputs}) and \textit{conclusions} (\textit{outputs}). We draw from disjoint collections of variables (written $\vara,\varb,\varc$, possiby sub- or superscripted) and covariables ($\covara,\covarb,\covarc$) to represent in- and outputs by labeled formulas $A^{\vara}$ and $A^{\covara}$. The latter combine into (antecedent )structures $\Gamma,\Delta$ and co(nsequent )structures $\Pi,\Sigma$, as specified by
\footnote{The reader eager to indulge in notational overloading may note that the symbols $\stimes,\splus$ suffice for representing each of the binary structural operations. E.g., $\Gamma\stimes\Delta$, $\Gamma\stimes\Sigma$ and $\Gamma\stimes\Delta$ are unambiguously recognized as $\Gamma\stimes\Delta$, $\Gamma\sos\Sigma$ and $\Pi\sobs\Gamma$ respectively.} 
\begin{center}
\begin{tabular}{lclr}
$\Gamma,\Delta$ & $::=$ & $A^{\vara}$ $|$ $(\Gamma\stimes\Delta)$ $|$ $(\Gamma\sos\Sigma)$ $|$ $(\Pi\sobs\Gamma)$  $|$ $\srdgc{\Pi}$ $|$ $\sldgc{\Sigma}$ & \ \ \ \ \ Structures \\
$\Pi,\Sigma$ & $::=$ & $A^{\covara}$ $|$ $(\Sigma\splus\Pi)$ $|$ $(\Pi\sbs\Gamma)$ $|$ $(\Delta\ssl\Pi)$ $|$ $\slgc{\Delta}$ $|$ $\srgc{\Gamma}$ & \ \ \ \ \ Costructures
\end{tabular}
\end{center}
The various constructors involved are seen as structural counterparts for the logical connectives via the the following translation tables:
\begin{center}
\begin{tabular}{|l|l||l|l||l|l|}\hline
 & $F(\cdot)$ & & $F(\cdot)$ & & $F(\cdot)$ \\ \hline
$A^{\vara}$ & $A$ & $A^{\covara}$ & $A$ & $\slgc{\Delta}$ & $\negl{F(\Delta)}$ \\
$\Gamma\stimes\Delta$ & $F(\Gamma)\lgtimes F(\Delta)$ & $\Sigma\splus\Pi$ & $F(\Pi)\lgplus F(\Sigma)$ & $\srgc{\Gamma}$ & $\negr{F(\Gamma)}$ \\
$\Pi\sobs\Gamma$ & $F(\Pi)\obs F(\Gamma)$ & $\Pi\sbs\Gamma$ & $F(\Gamma)\bs F(\Pi)$ & $\srdgc{\Pi}$ & $\conegr{F(\Pi)}$ \\
$\Gamma\sos\Sigma$ & $F(\Gamma)\os F(\Sigma)$ & $\Delta\ssl\Pi$ & $F(\Pi)/F(\Delta)$ & $\sldgc{\Sigma}$ & $\conegl{F(\Sigma)}$ \\ \hline
\end{tabular}
\end{center}
We have sided with display calculi (\cite{gore98}) in rejecting the standard practice of allowing only conjunction (in the antecedent) and disjunction (consequent) to be understood structurally. The association of types $\lsem A\rsem$ with formulae $A$ extends to a mapping of (co)structures into \textbf{LP}-contexts. In the base case, we stipulate\footnote{We assume each (co)variable of \textbf{LG} to have been uniquely associated with a variable of \textbf{LP}. The details of this correspondence are abstracted away from in our notation: the context is to differentiate between \textbf{LG}'s (co)variables and their \textbf{LP} counterparts.}
\begin{center}
\begin{tabular}{ccc}
$\lsem A^{\vara}\rsem=\left\{\begin{array}{rl}\{\lsem A\rsem^{\vara}\} & \textrm{if }\epsilon(A)=+ \\ \{\lnot\lsem A\rsem^{\vara}\} & \textrm{if }\epsilon(A)=-\end{array}\right.$ & \ \ \ \ \ &
$\lsem A^{\covara}\rsem=\left\{\begin{array}{rl}\{\lnot\lsem A\rsem^{\covara}\} & \textrm{if }\epsilon(A)=+ \\ \{\lsem A\rsem^{\covara}\} & \textrm{if }\epsilon(A)=-\end{array}\right.$
\end{tabular}
\end{center}
while structural connectives collapse into multiset union. The underlying intuition: inputs occupy the downward monotone arguments of an implication, while outputs instantiate the upward monotone ones (see also the entries for implications in Table \ref{polarities}).  

\subsubsection{The display property.}
We assert inequalities $F(\Gamma)\leq F(\Pi)$ through \textit{sequents} $\seq{\Gamma}{\Pi}{M}$ or $\seq{\Pi}{\Gamma}{M}$ (the relative ordering of $\Gamma,\Pi$ being irrelevant), where $\seqlpd{\sem{\Gamma},\sem{\Pi}}{M}{\result}$. Our use of structural counterparts for (co)implications scatters in- and outputs all over the sequent, as opposed to nicely partitioning them into (the yields of) the antecedent and consequent structures. Instead, the following \textit{display postulates}, mapping to (co)residuation and (co)Galois laws under $F$,  allow each input (output) to be displayed as the whole of the antecedent (consequent):
\begin{center}
\begin{tabular}{ccccccc}
$\infer={\seq{\Pi}{\Gamma}{M}}{\seq{\Gamma}{\Pi}{M}}$ & \ \ \ \ \ &
$\infer={\seq{\srgc{\Gamma}}{\Delta}{M}}{\seq{\Gamma}{\slgc{\Delta}}{M}}$ & \ \ \ \ \ &
$\infer={\seq{\Sigma}{\srdgc{\Pi}}{M}}{\seq{\sldgc{\Sigma}}{\Pi}{M}}$ 
\end{tabular}
\end{center}
\begin{center}
\begin{tabular}{ccccccc}
$\infer={\seq{\Gamma}{\Delta\ssl\Pi}{M}}{\seq{\Gamma\stimes\Delta}{\Pi}{M}}$ & \ \ \ \ \ &
$\infer={\seq{\Pi\sbs\Gamma}{\Delta}{M}}{\seq{\Pi}{\Gamma\stimes\Delta}{M}}$ & \ \ \ \ \ &
$\infer={\seq{\Gamma\sos\Sigma}{\Pi}{M}}{\seq{\Gamma}{\Sigma\splus\Pi}{M}}$ & \ \ \ \ \ &
$\infer={\seq{\Sigma}{\Pi\sobs\Gamma}{M}}{\seq{\Sigma\splus\Pi}{\Gamma}{M}}$
\end{tabular}
\end{center}
Sequents $\seq{\Gamma}{\Pi}{M}$ and $\seq{\Delta}{\Sigma}{M}$ are declared \textit{display-equivalent} iff they are interderivable using only the display postulates, a fact we often abbreviate
\begin{center}
$\infer=[dp]{\seq{\Delta}{\Sigma}{M}}{\seq{\Gamma}{\Pi}{M}}$
\end{center}
The \textit{display property} now reads: for any input $A^{\vara}$ appearing in $\seq{\Gamma}{\Pi}{M}$, there exists some $\Sigma$ such that $\seq{\Sigma}{A^{\vara}}{M}$ and $\seq{\Gamma}{\Pi}{M}$ are display-equivalent, and similarly for outputs. 

\subsubsection{Focused proof search.} 
We shall allow a displayed hypothesis or conclusion to inhabit the righthand zone of $\vdash$, named the \textit{stoup}, after \cite{girard91}. Thus, $\seqs{\Gamma}{M}{A}$ and $\seqs{\Pi}{M}{A}$ are sequents, provided
\begin{center}
\begin{tabular}{rclr}
$\seqlpd{\sem{\Gamma}}{M}{\sem{A}}$ & and & $\seqlpd{\sem{\Pi}}{M}{\lnot\sem{A}}$ & if $\epsilon(A)=+$ \\
$\seqlpd{\sem{\Gamma}}{M}{\lnot\sem{A}}$ & and & $\seqlpd{\sem{\Pi}}{M}{\sem{A}}$ & if $\epsilon(A)=-$
\end{tabular}
\end{center}
The presence of a stoup implements Andreoli's concept of \textit{focused proof search} (\cite{andreoli92}). That is, when working one's way backwards from the desired conclusion to the premises, one commits to the contents of the stoup (the \textit{focus}) as the main formula: the only (logical) inference rules deriving a sequent $\seqs{\Gamma}{M}{A}$ or $\seqs{\Pi}{M}{A}$ are those introducing $A$, and focus propagates to the subformulas of $A$ appearing in the premises. The pay-off: a reduction of the search space.

We need structural rules for managing the contents of the stoup. As will be argued below, focusing is relevant only for the negative inputs and positive outputs. Thus, we have \textit{decision} rules for moving the latter inside the stoup, and \textit{reaction} rules for taking their duals out:
\begin{center}\small
\begin{tabular}{ccccccc}
$\infer[D^{\bullet}]{\seq{\Pi}{K^{\vara}}{(\vara \ M)}}{\seqs{\Pi}{M}{K}}$ & \ &
$\infer[D^{\circ}]{\seq{\Gamma}{P^{\covara}}{(\covara \ M)}}{\seqs{\Gamma}{M}{P}}$ & \ &
$\infer[R^{\bullet}]{\seqs{\Pi}{\lambda\vara^{\sem{P}}M}{P}}{\seq{\Pi}{P^{\vara}}{M}}$ & \ &
$\infer[R^{\circ}]{\seqs{\Gamma}{\lambda\covara^{\sem{K}}M}{K}}{\seq{\Gamma}{K^{\covara}}{M}}$
\end{tabular}\normalsize
\end{center}

\subsubsection{Logical inferences.} 
Each connective has its meaning defined by two types of rules: one infering it from its structural countepart (affecting the positive inputs and negative outputs) and one introducing it \textit{alongside} its structural counterpart via monotonicity (targeting negative inputs and positive outputs). In reference to Smullyan's unified notation (\cite{smullyan}), we speak of rules of type $\alpha$ and $\beta$ respectively. The former preserve provability of the conclusion in their premises and hence come for free in proof search, meaning we may apply them nondeterministically. In contrast, the order in which the available $\beta$-type rules are tried may well affect success. Not all of the non-determinism involved is meaningful, however, as witnessed by the trivial permutations of $\beta$-type rules involving disjoint active formulas. Their domain of influence we therefore restrict to the stoup.\footnote{While reducing the search space, it is not immediate that completeness w.r.t. provability in \textbf{LG} is preserved. We return to this issue at the end of this section.} It follows that we may interpret term construction under $(\alpha)$ and $(\beta)$ by the \textbf{LP}-inferences $(\lgtimes E)$ and $(\lgtimes I)$ respectively. Using the meta-variables
\begin{center}
\begin{tabular}{lrl}
 & $\nvara,\nvarb,\nvarc$ & \ \ for variables and covariables \\
and \ & $\Theta,\Theta_1,\Theta_2$ & \ \ for antecedent and consequent structures,
\end{tabular}
\end{center}
we may formalize the above intuitions by the following tables and rule schemata:
\begin{center}
\begin{tabular}{ccc}
\begin{tabular}{|l|l|l|l|} \hline
$\Theta,\alpha^{\nvara}$ & $\alpha_1^{\nvarb}$ & $\alpha_2^{\nvarc}$ & $*$ \\ \hline
$\Pi,A\lgtimes B^{\vara}$ & $A^{\varb}$ & $B^{\varc}$ & $\stimes$ \\
$\Gamma,A/B^{\covara}$ & $B^{\varb}$ & $A^{\covarc}$ & $\ssl$ \\
$\Gamma,B\bs A^{\covara}$ & $A^{\covarb}$ & $B^{\varc}$ & $\sbs$ \\
$\Gamma,A\lgplus B^{\covara}$ & $B^{\covarb}$ & $A^{\covarc}$ & $\splus$ \\
$\Pi,B\obs A^{\vara}$ & $B^{\covarb}$ & $A^{\varc}$ & $\sobs$ \\
$\Pi,A\os B^{\vara}$ & $A^{\varb}$ & $B^{\covarc}$ & $\sos$ \\ \hline
\end{tabular} & \ \ \ \ \ &
\begin{tabular}{|l|l|l|l|} \hline
$\beta$ & $\Theta_1,\beta_1$ & $\Theta_2,\beta_2$ & $*$ \\ \hline
$A\lgtimes B$ & $\Gamma,A$ & $\Delta,B$ & $\stimes$ \\
$A/B$ & $\Delta,B$ & $\Pi,A$ & $\ssl$ \\
$B\bs A$ & $\Pi,A$ & $\Delta,B$ & $\sbs$ \\
$A\lgplus B$ & $\Sigma,B$ & $\Pi,A$ & $\splus$ \\
$B\obs A$ & $\Sigma,B$ & $\Gamma,A$ & $\sobs$ \\
$A\os B$ & $\Gamma,A$ & $\Sigma,B$ & $\sos$ \\ \hline
\end{tabular} \\ \\
$\infer[\alpha]{\seq{\Theta}{\alpha^{\nvara}}{\casefusion{\nvara}{\nvarb}{\nvarc}{M}}}{\seq{\Theta}{\alpha_1^{\nvarb}\mathbin{*}\alpha_2^{\nvarc}}{M}}$ & &
$\infer[\beta]{\seqs{\Theta_1*\Theta_2}{\fusionpair{M}{N}}{\beta}}{
\seqs{\Theta_1}{M}{\beta_1} &
\seqs{\Theta_2}{N}{\beta_2}}$
\end{tabular}
\end{center}
and for the unary connectives (overloading the $\alpha,\beta$ notation):
\begin{figure}
\begin{center}
\begin{tabular}{ccc}
$\begin{array}{c}\infer[\alpha]{\seq{\Pi}{P\lgtimes Q^{\vara}}{\casefusion{\vara}{\varb^{\sem{P}}}}{\varc^{\sem{Q}}}{M}}{\seq{\Pi}{P^{\varb}\stimes Q^{\varc}}{M}}\end{array}$ & \ & 
$\begin{array}{c}\infer[\beta]{\seqs{\Gamma\stimes\Delta}{\fusionpair{M}{N}}{P\lgtimes Q}}{
\seqs{\Gamma}{M}{P} &
\seqs{\Delta}{N}{Q}}\end{array}$ \\ \\

$\begin{array}{c}\infer[\alpha]{\seq{\Pi}{P\lgtimes L^{\vara}}{\casefusion{\vara}{\varb^{\sem{P}}}}{\varc^{\lnot\sem{L}}}{M}}{\seq{\Pi}{P^{\varb}\stimes L^{\varc}}{M}}\end{array}$ & \ &
$\begin{array}{c}\infer[\beta]{\seqs{\Gamma\stimes\Delta}{\fusionpair{M}{\lambda\covarc^{\sem{L}}N}}{P\lgtimes L}}{
\seqs{\Gamma}{M}{P} &
\infer[R^{\splus}]{\seqs{\Delta}{\lambda\covarc^{\sem{L}}M}{L}}{\seq{\Delta}{L^{\covarc}}{N}}}\end{array}$ \\ \\

$\begin{array}{c}\infer[\alpha]{\seq{\Pi}{K\lgtimes Q^{\vara}}{\casefusion{\vara}{\varb^{\lnot\sem{K}}}}{\varc^{\sem{Q}}}{M}}{\seq{\Pi}{K^{\varb}\stimes Q^{\varc}}{M}}\end{array}$ & \ &
$\begin{array}{c}\infer[\beta]{\seqs{\Gamma\stimes\Delta}{\fusionpair{\lambda\covarb^{\sem{K}}M}{N}}{K\lgtimes Q}}{
\infer[R^{\splus}]{\seqs{\Gamma}{\lambda\covarb^{\sem{K}}M}{K}}{
\seq{\Gamma}{K^{\covarb}}{M}} &
\seqs{\Delta}{N}{Q}}\end{array}$ \\ \\

$\begin{array}{c}\infer[\alpha]{\seq{\Pi}{K\lgtimes L^{\vara}}{\casefusion{\vara}{\varb^{\lnot\sem{K}}}}{\varc^{\lnot\sem{L}}}{M}}{\seq{\Pi}{K^{\varb}\stimes L^{\varc}}{M}}\end{array}$ & \ &
$\begin{array}{c}\infer[\beta]{\seqs{\Gamma\stimes\Delta}{\fusionpair{\lambda\covarb^{\sem{K}}.M}{\lambda\covarc^{\sem{L}}N}}{K\lgtimes L}}{
\infer[R^{\splus}]{\seqs{\Gamma}{\lambda\covarb^{\sem{K}}M}{K}}{
\seq{\Gamma}{K^{\covarb}}{M}} &
\infer[R^{\splus}]{\seqs{\Delta}{\lambda\covarc^{\sem{L}}M}{L}}{\seq{\Delta}{L^{\covarc}}{N}}}\end{array}$

\end{tabular}
\end{center}
\caption{Checking all possible instantiations of $(\alpha),(\beta)$ for $\lgtimes$. We also mention obligatory reactions ($R^{\stimes},R^{\splus}$) in the premises of $(\beta)$.}
\label{rules_tensor}
\end{figure}
\begin{center}
\begin{tabular}{ccccc}
\begin{tabular}{|l|l|l|} \hline
$\Theta,\alpha^{\nvara}$ & $\alpha_1^{\nvarb}$ & $\cdot^*$ \\ \hline
$\Gamma,\negl{A}^{\covara}$ & $A^{\varb}$ & $\slgc{\cdot}$ \\ 
$\Gamma,\negr{A}^{\covara}$ & $A^{\varb}$ & $\srgc{\cdot}$ \\
$\Pi,\conegr{A}^{\vara}$ & $A^{\covarb}$ & $\srdgc{\cdot}$ \\
$\Pi,\conegl{A}^{\vara}$ & $A^{\covarb}$ & $\sldgc{\cdot}$ \\ \hline
\end{tabular} & \ \ \ \ \ &
\begin{tabular}{|l|l|l|} \hline
$\beta$ & $\Theta,\beta_1$ & $\cdot^*$ \\ \hline
$\negl{A}$ & $\Delta,A$ & $\slgc{\cdot}$ \\
$\negr{A}$ & $\Delta,A$ & $\srgc{\cdot}$ \\
$\conegr{A}$ & $\Sigma,A$ & $\srdgc{\cdot}$ \\
$\conegl{A}$ & $\Sigma,A$ & $\sldgc{\cdot}$ \\ \hline
\end{tabular} & \ \ \ \ \ &
$\begin{array}{c}\infer[\alpha]{\seq{\Theta}{\alpha^{\nvara}}{M[\nvara/\nvarb]}}{\seq{\Theta}{\alpha_1^{\nvarb *}}{M}} \\ \\
\infer[\beta]{\seqs{\Theta^*}{M}{\beta}}{\seqs{\Theta}{M}{\beta_1}}\end{array}$
\end{tabular}
\end{center}
Well-definedness is established through a case-by-case analysis. To illustrate, Figure \ref{rules_tensor} checks all possible instantiations of $(\alpha),(\beta)$ for $\lgtimes$. Finally, assigning positive bias to atoms implies Axioms have their conclusion placed in focus:
\begin{center}
$\infer[Ax]{\seqs{p^{\vara}}{\vara}{p}}{}$
\end{center}
\subsubsection{Soundness and completeness.} In what preceded, we have already informally motivated soundness w.r.t. $\S$1's algebraic formulation of \textbf{LG}. Completeness is demonstrated in a companion paper under preparation. Roughly, we define a syntactic phase model wherein every truth is associated with a Cut-free focused proof, similar to \cite{okada02}.
\begin{figure}[h]
\begin{center}
\begin{tabular}{ccccccc}
$\infer[Ax]{\seqs{p^{\vara}}{\vara}{p}}{}$ & \ \ \ \ \ &
$\infer={\seq{\Pi}{\Gamma}{M}}{\seq{\Gamma}{\Pi}{M}}$ & \ \ \ \ \ &
$\infer={\seq{\Gamma}{\Delta\ssl\Pi}{M}}{\seq{\Gamma\stimes\Delta}{\Pi}{M}}$ \\ \\
$\infer={\seq{\Pi\sbs\Gamma}{\Delta}{M}}{\seq{\Pi}{\Gamma\stimes\Delta}{M}}$ & \ \ \ \ \ &
$\infer={\seq{\Gamma\sos\Sigma}{\Pi}{M}}{\seq{\Gamma}{\Sigma\splus\Pi}{M}}$ & \ \ \ \ \ &
$\infer={\seq{\Sigma}{\Pi\sobs\Gamma}{M}}{\seq{\Sigma\splus\Pi}{\Gamma}{M}}$ \\ \\
\end{tabular} \\
\begin{tabular}{ccc}
$\infer[D^{\bullet}]{\seq{\Pi}{K^{\vara}}{\lambda\vara^{\sem{K}}M}}{\seqs{\Pi}{M}{K}}$ & \ \ \ \ \ &
$\infer[D^{\circ}]{\seq{\Gamma}{P^{\covara}}{\lambda\covara^{\sem{P}}M}}{\seqs{\Gamma}{M}{P}}$ \\ \\
$\infer[R^{\bullet}]{\seqs{\Pi}{(\vara \ M)}{P}}{\seq{\Pi}{P^{\vara}}{M}}$ & \ \ \ \ \ &
$\infer[R^{\circ}]{\seqs{\Gamma}{(\covara \ M)}{K}}{\seq{\Gamma}{K^{\covara}}{M}}$ \\ \\
$\infer[\alpha]{\seq{\Theta}{\alpha^{\nvara}}{\casefusion{\nvara}{\nvarb}{\nvarc}{M}}}{\seq{\Theta}{\alpha_1^{\nvarb}\mathbin{*}\alpha_2^{\nvarc}}{M}}$ & \ \ \ \ \ &
$\infer[\beta]{\seqs{\Theta_1*\Theta_2}{\fusionpair{M}{N}}{\beta}}{
\seqs{\Theta_1}{M}{\beta_1} &
\seqs{\Theta_2}{N}{\beta_2}}$ \\ \\
$\infer[\alpha]{\seq{\Theta}{\alpha^{\nvara}}{M[\nvara/\nvarb]}}{\seq{\Theta}{\alpha_1^{\nvarb *}}{M}}$ & \ \ \ \ \ &
$\infer[\beta]{\seqs{\Theta^*}{M}{\beta}}{\seqs{\Theta}{M}{\beta_1}}$
\end{tabular}
\end{center}
\caption{An overview of the term-labeled sequent calculus for \textbf{LG}. An easy induction shows that terms labeling derivable sequents are in ($\beta$-)normal form.}
\label{lgfoc_rules}
\end{figure}

\section{Case analysis: Extraction}
We illustrate our semantics of $\S$2 alongside an analysis of extraction phenomena. Syntactically, their treatment in \textbf{NL} necessitates controlled associativity and commutativity (\cite{moortgat97}). Kurtonina and Moortgat (K$\&$M, \cite{kurtoninamoortgat}) ask whether the same results are obtainable in \textbf{LG} from having $\lgtimes$ and $\lgplus$ interact through linear distributivity. We work out the details of such an approach, after first having pointed out a flaw in an alternative proposal by K$\&$M. As illustration, we work out the derivational and lexical semantics of several sample sentences.

\subsubsection{The good.}
We first consider a case that already works fine in \textbf{NL}. The following complex noun demonstrates extraction out of (subordinate) subject position:
\begin{itemize}
\item[(1)]\label{data_1} (the) mathematician who founded intuitionism
\end{itemize}
We analyze (1) into a binary branching tree, categorizing the words as follows:
\begin{center}
\begin{tabular}{cccc}
mathematician \ & $[$ who & \ $[$ invented \ & \ intuitionism \ $]]$ \\
$n$ & \ $(n\bs n)/(\np\bs s)$ \ & $(\np\bs s)/\np$ & $np$ \\ 
\end{tabular}
\end{center}
employing atoms $s$ (categorizing sentences), $\np$ (noun phrases) and $n$ (nouns). Note the directionality in the category $\np\bs s$ assigned to the gapped clause (as selected for by \textit{who}), seeing as the $\np$ gap occurs in a left branch.  Figure \ref{deriv_relative} demonstrates (1), with bracketing as indicated above, to be categorizable by $n$, referring to the 'macro' from Figure \ref{deriv_sample} for deriving transitive clauses. 
\begin{figure}[h]
\begin{center}
$\infer=[dp]{\seq{\np^{\vara}\stimes((\np\bs s)/\np^{\varb}\stimes\np^{\varc})}{s^{\covarc}}{(\varb \ \fusionpair{\varc}{\fusionpair{\lambda u(\covarc \ u)}{\vara}})}}{
\infer[D^{\bullet}]{\seq{\np^{\varc}\ssl(s^{\covarc}\sbs\np^{\vara})}{(\np\bs s)/\np^{\varc}}{(\varb \ \fusionpair{\varc}{\fusionpair{\lambda u(\covarc \ u)}{\vara}})}}{
\infer[\beta]{\seqs{\np^{\varc}\ssl(s^{\covarc}\sbs\np^{\vara})}{\fusionpair{\varc}{\fusionpair{\lambda u(\covarc \ u)}{\vara}}}{(\np\bs s)/\np}}{
\infer[\beta]{\seqs{s^{\covarc}\sbs\np^{\vara}}{\fusionpair{\lambda u(\covarc \ u)}{\vara}}{\np\bs s}}{
\infer[R^{\bullet}]{\seqs{s^{\covarc}}{\lambda u(\covarc \ u)}{s}}{
\infer[D^{\circ}]{\seq{s^{\covarc}}{s^u}{(\covarc \ u)}}{
\infer[Ax]{\seqs{s^u}{u}{s}}{}}} &
\infer[Ax]{\seqs{\np^{\vara}}{\vara}{\np}}{}} &
\infer[Ax]{\seqs{\np^{\varc}}{\varc}{\np}}{}}}}$
\end{center}
\caption{Derivation of a transitive clause in an SVO language.}
\label{deriv_sample}
\end{figure}

We proceed with a specification of the lexical semantics. Linearity no longer applies at this stage, as our means of referring to the world around us is not so restricted. Thus, we allow access to the full repertoire of the simply-typed $\lambda$-calculus, augmented with logical constants for the propositional connectives. Concretely, lexical denotations are built over types
\begin{center}
$\tau,\sigma$ $::=$ $e$ $|$ $t$ $|$ $(\tau\times\sigma)$ $|$ $(\tau\rightarrow\sigma)$
\end{center}
where $e,t$ interpret (a fixed set of) entities and (Boolean) truth values respectively. The linear types and terms of $\S$2 carry over straightforwardly: interpret $\bot$, $\tau\lgtimes\sigma$ and $\lnot\tau$ by $t$, $\tau\times\sigma$ and $\tau\rightarrow t$, with terms $\fusionpair{M}{N}$ and $\casefusion{N}{x}{y}{M}$ being replaced by pairs $\langle M,N\rangle$ and projections $M[\pi_1(N)/x,\pi_2(N)/y]$. The remaining atoms $s$, $\np$ and $n$ we interpret by $t$ (sentences denote truth values), $e$ (noun phrases denote entities) and $e\rightarrow t$ (nouns denote first-order properties) respectively.  Abbreviating $\lambda\vara^{\tau\times\sigma}M[\pi_1(x)/y,\pi_2(x)/z]$ by $\lambda\langle y,z\rangle M$ and types $\tau\rightarrow t$ by $\lnot\tau$, the linear terms of $\S$2 remain practically unchanged. For instance, delinearization of the term found in Figure \ref{deriv_relative} for (1) gives
\begin{center}
$(w \ \langle\lambda\langle\covarb,b\rangle(f \ \langle i,\langle\lambda z(\covarb \ z),b\rangle\rangle),\langle\lambda y(\covarc \ y),m\rangle\rangle)$
\end{center}
the free variables $w$, $f$, $i$ and $m$ ranging over the denotations of \textit{who}, \textit{founded}, \textit{intuitionism} and \textit{mathematician}. Since words act as inputs, those categorized $P$ ($K$) are interpreted by a closed term $M$ of type $\sem{P}$ ($\lnot\sem{K}$). These remarks motivate the following lexical entries, conveniently written as nonlogical axioms:
\begin{center}
\begin{tabular}{rcl}
mathematician & $\vdash$ & $\textsc{mathematician}:n$ \\
who & $\vdash$ & $\lambda\langle Q,\langle\covarc,P\rangle\rangle(\covarc \ \lambda x((P \ x)\land(Q \ \langle\lambda pp,x\rangle))):(n\bs n)/(\np\bs s)$ \\
founded & $\vdash$ & $\lambda\langle y,\langle q,x\rangle\rangle(q \ ((\textsc{founded} \ y) \ x)):(\np\bs s)/\np$ \\
intuitionism & $\vdash$ & $\textsc{intuitionism}:\np$
\end{tabular}
\end{center}
We applied the familiar trick of switching fonts to abstract away from certain interpretations, yielding constants \textsc{mathematician} (type $\lnot e$), \textsc{founded} ($e\rightarrow\lnot e$) and \textsc{intuitionism} ($e$). If we take the nonlogical axiom perspective seriously, lexical substitution proceeds via Cut. Simplifying, we directly substitute terms for the free variables $m,w,f$ and $i$, yielding, after $\beta$-reduction, the term  (with free variable $\gamma$ corresponding to the assigned category $n$)
\begin{center}
$(\gamma \ \lambda x((\textsc{mathematician} \ x)\land((\textsc{founded} \ \textsc{intuitionism}) \ x))),$
\end{center}

\subsubsection{The bad.} Cases of non-subject extraction are illustrated in (2) and (3) below:
\begin{itemize}
\item[(2)]\label{data_2} (the) law that Brouwer rejected
\item[(3)]\label{data_3} (the) mathematician whom TNT pictured on a post stamp
\end{itemize}
While tempting to categorize \textit{that} and \textit{whom} by $(n\bs n)/(s/\np)$ (noticing the gap now occurs in a right branch), we find that  derivability of (2) then necessitates rebracketing (mentioning also the dual concept for reasons of symmetry):
\begin{center}
\begin{tabular}{ccc}
$(A\lgtimes B)\lgtimes C\leq A\lgtimes(B\lgtimes C)$ & \ \ \ \ \ & $A\lgplus(B\lgplus C)\leq(A\lgplus B)\lgplus C$ \\
$A\lgtimes(B\lgtimes C)\leq (A\lgtimes B)\lgtimes C$ & \ \ \ \ \ & $(A\lgplus B)\lgplus C\leq A\lgplus(B\lgplus C)$
\end{tabular}
\end{center}
Worse yet, (3) requires (weak) commutativity for its derivability:
\begin{center}
\begin{tabular}{ccc}
$(A\lgtimes B)\lgtimes C\leq(A\lgtimes C)\lgtimes B$ & \ \ \ \ \ & $(A\lgplus C)\lgplus B\leq(A\lgplus B)\lgplus C$ \\
$A\lgtimes(B\lgtimes C)\leq B\lgtimes(A\lgtimes C)$ & \ \ \ \ \ & $B\lgplus(A\lgplus C)\leq A\lgplus(B\lgplus C)$
\end{tabular}
\end{center}
Said principles, however, contradict the resource sensitive nature of linguistic reality. Kurtonina and Moortgat (K$\&$M, \cite{kurtoninamoortgat}), working in \textbf{LG}, questioned the viability of a different solution: revise the categorization of \textit{whom} such that recourse need be made only to linear distributivity of $\lgtimes$ over $\lgplus$ (or \textit{mixed} associativity and commutativity, if you will):
\begin{center}
\begin{tabular}{ccc}
$(A\lgplus B)\lgtimes C\leq A\lgplus(B\lgtimes C)$ & \ \ \ \ \ & $(A\lgplus B)\lgtimes C\leq(A\lgtimes C)\lgplus B$ \\
$A\lgtimes(B\lgplus C)\leq(A\lgtimes B)\lgplus C$ & \ \ \ \ \ & $A\lgtimes(B\lgplus C)\leq B\lgplus(A\lgtimes C)$
\end{tabular}
\end{center}
As observed by Moortgat (using a slightly different syntax), the presence of (co)implications allows a presentation in the following rule format:
\begin{center}
\begin{tabular}{ccc}
$\infer[(\obs,/)]{\seq{\Gamma\stimes\Delta}{\Sigma\splus\Pi}{M}}{\seq{\Pi\sobs\Gamma}{\Delta\ssl\Sigma}{M}}$ & \ \ \ \ \ &
$\infer[(\os,\bs)]{\seq{\Gamma\stimes\Delta}{\Sigma\splus\Pi}{M}}{\seq{\Delta\sos\Sigma}{\Pi\sbs\Gamma}{M}}$ \\ \\
$\infer[(\obs,\bs)]{\seq{\Gamma\stimes\Delta}{\Sigma\splus\Pi}{M}}{\seq{\Pi\sobs\Delta}{\Sigma\sbs\Gamma}{M}}$ & &
$\infer[(\os,/)]{\seq{\Gamma\stimes\Delta}{\Sigma\splus\Pi}{M}}{\seq{\Gamma\sobs\Sigma}{\Delta\ssl\Pi}{M}}$
\end{tabular}
\end{center}
K$\&$M suggested in particular to categorize \textit{whom} by $(n\bs n)/((s\os s)\lgplus(s/\np))$. However, their analysis assumes $(\obs,/)$, $(\os,\bs)$, $(\obs,\bs)$ and $(\os,/)$ to be \textit{invertible}, thereby seriously compromising the resource sensitivity of \textbf{LG}, as illustrated by the derivable inferences of Figure \ref{no_14} (and similar ones under $\cdot^{\infty}$).

\subsubsection{The analysis.} We propose a solution to K$\&$M's challenge by categorizing \textit{whom} using not only the (co)residuated families of connectives, but also the Galois connected pair $\negl{\cdot},\negr{\cdot}$. In particular, we have in mind the following lexicon for (2):
\begin{center}
\begin{tabular}{rcl}
law & $\vdash$ & $\textsc{law}:n$ \\
that & $\vdash$ & $\lambda\langle Q,\langle\covarc,P\rangle\rangle(\covarc \ \lambda x((P \ x)\land(Q \ \langle\lambda pp,x\rangle))):(n\bs n)/(s\lgplus\negl{np})$ \\
Brouwer & $\vdash$ & $\textsc{brouwer}:\np$ \\
rejected & $\vdash$ & $\lambda\langle y,\langle q,x\rangle\rangle(q \ ((\textsc{rejected} \ y) \ x)):(\np\bs s)/\np$
\end{tabular}
\end{center}
employing constants \textsc{law}, \textsc{brouwer} and \textsc{rejected} of types $\lnot e$, $e$ and $e\rightarrow\lnot e$. Note the formula $s\lgplus\negl{\np}$ (selected for by \textit{that}) categorizing the gapped clause; had the gap occurred in a left branch, we would have used $\negr{\np}\lgplus s$ instead. Figure \ref{deriv_relative} gives the derivation. Lexical substitution and $\beta$-reduction yield
\begin{figure}
\begin{center}
\begin{tabular}{ccc}
$\infer=[(\os,\bs)]{\seq{\Gamma_1\stimes(\Gamma_2\stimes\Gamma_3)}{\Sigma\splus\Pi}{M}}{
\infer=[dp]{\seq{(\Gamma_2\stimes\Gamma_3)\sos\Sigma}{\Pi\sbs\Gamma_1}{M}}{
\infer=[(\obs,\bs)]{\seq{\Gamma_2\stimes\Gamma_3}{\Sigma\splus(\Pi\sbs\Gamma_1)}{M}}{
\infer=[dp]{\seq{(\Pi\sbs\Gamma_1)\sobs\Gamma_3}{\Sigma\sbs\Gamma_2}{M}}{
\infer=[(\os,\bs)]{\seq{\Gamma_3\sos(\Sigma\sbs\Gamma_2)}{\Pi\sos\Gamma_1}{M}}{
\infer=[dp]{\seq{\Gamma_1\stimes\Gamma_3}{(\Sigma\sbs\Gamma_2)\splus\Pi}{M}}{
\infer=[(\obs,\bs)]{\seq{\Pi\sobs(\Gamma_1\stimes\Gamma_3)}{\Sigma\sbs\Gamma_2}{M}}{
\seq{\Gamma_2\stimes(\Gamma_1\stimes\Gamma_3)}{\Sigma\splus\Pi}{M}}}}}}}}$ & \ \ \ \ &

$\infer=[(\os,\bs)]{\seq{\Gamma_1\stimes(\Gamma_2\stimes\Gamma_3)}{\Sigma\splus\Pi}{M}}{
\infer=[dp]{\seq{(\Gamma_2\stimes\Gamma_3)\sos\Sigma}{\Pi\sbs\Gamma_1}{M}}{
\infer=[(\obs,/)]{\seq{\Gamma_2\stimes\Gamma_3}{\Sigma\splus(\Pi\sbs\Gamma_1)}{M}}{
\infer=[dp]{\seq{(\Pi\sbs\Gamma_1)\sobs\Gamma_2}{\Gamma_3\ssl\Sigma}{M}}{
\infer=[(\obs,\bs)]{\seq{\Gamma_2\sos(\Gamma_3\ssl\Sigma)}{\Pi\sbs\Gamma_1}{M}}{
\infer=[dp]{\seq{\Gamma_1\stimes\Gamma_2}{(\Gamma_3\ssl\Sigma)\splus\Pi}{M}}{
\infer=[(\obs,/)]{\seq{\Pi\sobs(\Gamma_1\stimes\Gamma_2)}{\Gamma_3\ssl\Sigma}{M}}{
\seq{(\Gamma_1\stimes\Gamma_2)\stimes\Gamma_3}{\Sigma\splus\Pi}{M}}}}}}}}$
\end{tabular}
\end{center}
\caption{Illustrating the structural collapse induced by making $(\os,/)$, $(\os,\bs)$, $(\obs,/)$ and $(\obs,\bs)$ invertible.}
\label{no_14}
\end{figure}
\begin{center}
$(\gamma \ \lambda x((\textsc{law} \ x)\land((\textsc{rejected} \ x) \ \textsc{brouwer})))$
\end{center}
Like K$\&$M, we have, in Figure \ref{deriv_relative}, not relied exclusively on linear distributivity: the (dual) Galois connected pairs now go 'halfway De Morgan', as explicated by the following three equivalent groups of axioms 
\begin{center}
\begin{tabular}{ccccc}
$\conegr{(A\lgtimes B)}\leq\negl{B}\lgplus\negl{A}$ & \ \ \ \ & $A\os\negl{B}\leq A\lgtimes B$ & \ \ \ \ & $A/B\leq A\lgplus\negl{B}$ \\
$\conegl{(A\lgtimes B)}\leq\negr{B}\lgplus\negr{A}$ & \ \ \ \ & $\negr{B}\obs A\leq B\lgtimes A$ & \ \ \ \ & $B\bs A\leq\negr{B}\lgplus A$ \\
$\conegr{A}\lgtimes\conegr{B}\leq\negl{(B\lgplus A)}$ & & $B\lgplus A\leq\conegr{B}\bs A$ & & $\conegr{B}\lgtimes A\leq B\obs A$ \\
$\conegl{A}\lgtimes\conegl{B}\leq\negr{(B\lgplus A)}$ & & $B\lgplus A\leq B/\conegl{A}$ & & $A\lgtimes\conegl{B}\leq A\os B$
\end{tabular}
\end{center}
Note their independence of their converses (i.e., with $\leq$ turned around). The following equivalent presentation in rule format is adapted from \cite{moortgat10}:
\begin{center}
\begin{tabular}{ccc}
$\infer[(\os,\negl{\cdot})]{\seq{\Gamma\sos\Pi}{\slgc{\Delta}}{M}}{\seq{\Gamma\stimes\Delta}{\Pi}{M}}$ & \ \ \ \ \ &
$\infer[(\obs,\negr{\cdot})]{\seq{\Pi\sobs\Delta}{\srgc{\Gamma}}{M}}{\seq{\Gamma\stimes\Delta}{\Pi}{M}}$ \\ \\
$\infer[(\os,\negr{\cdot})]{\seq{\Delta\sos\Pi}{\srgc{\Gamma}}{M}}{\seq{\Gamma\stimes\Delta}{\Pi}{M}}$ & &
$\infer[(\obs,\negl{\cdot})]{\seq{\Pi\sobs\Gamma}{\slgc{\Delta}}{M}}{\seq{\Gamma\stimes\Delta}{\Pi}{M}}$
\end{tabular}
\end{center}
The intuition behind our analysis is as follows. If we were to also adopt the converses of the above De Morgan axioms (turning $\leq$ around), same-sort associativity and weak commutativity would find equivalent presentations as
\begin{center}
\begin{tabular}{ccc}
$(A\lgtimes B)\os\negl{C}\leq A\lgtimes(B\os\negl{C})$ & \ \ \ \ \ & $(A\lgtimes B)\os\negl{C}\leq(A\os\negl{C})\lgtimes B$ \\
$\negr{A}\obs(B\lgtimes C)\leq (\negr{A}\obs B)\lgtimes C$ & \ \ \ \ \ & $\negr{A}\obs(B\lgtimes C)\leq B\lgtimes(\negr{A}\obs C)$
\end{tabular}
\end{center}
Going only halfway with De Morgan, however, the above inferences remain derivable (by virtue of linear distributivity) and useful (by composing with $A\os\negl{B}\leq A\lgtimes B$ and $\negr{B}\obs A\leq B\lgtimes A$), but without inducing a collapse. Indeed, none of the derivabilities of Figure \ref{no_14} carry over, and neither do the variations 
\begin{center}
\begin{tabular}{ccc}
$\infer={\seq{\Gamma_1\stimes(\Gamma_2\stimes\Gamma_3)}{\Sigma\splus\srgc{\Pi}}{M}}{\seq{(\Gamma_1\stimes\Gamma_2)\stimes\Gamma_3}{\Sigma\splus\srgc{\Pi}}{M}}$ & \ \ \ \ \ &
$\infer={\seq{\Gamma_1\stimes(\Gamma_2\stimes\Gamma_3)}{\Sigma\splus\srgc{\Pi}}{M}}{\seq{\Gamma_2\stimes(\Gamma_1\stimes\Gamma_3)}{\Sigma\splus\srgc{\Pi}}{M}}$
\end{tabular}
\end{center}
as an exhaustive exploration of the search space will tell, noting we need only consider structural rules. 

By virtue of the mixed commutativity involved in some of the linear distributivity postulates, it should be clear our formula $(n\bs n)/(s\lgplus\negl{\np})$ for \textit{that} in (2) also applies to \textit{whom} in (3), the latter example involving non-peripheral extraction. For reasons of space, we leave its analysis as an exercise.

\begin{sidewaysfigure}
\begin{center}
\begin{tabular}{c}
$\infer=[dp]{\seq{n^m\stimes((n\bs n)/(\np\bs s)^w\stimes((\np\bs s)/\np^f\stimes\np^i))}{n^{\covarc}}{(w \ \fusionpair{\lambda\fusionpair{\covarb}{b}(f \ \fusionpair{i}{\fusionpair{\lambda z(\covarb \ z)}{b}})}{\fusionpair{\lambda y(\covarc \ y)}{m}})}}{
\infer[D^{\bullet}]{\seq{((\np\bs s)/\np^f\stimes\np^i)\ssl(n^{\covarc}\sbs n^m)}{(n\bs n)/(\np\bs s)^w}{(w \ \fusionpair{\lambda\fusionpair{\covarb}{b}(f \ \fusionpair{i}{\fusionpair{\lambda z(\covarb \ z)}{b}})}{\fusionpair{\lambda y(\covarc \ y)}{m}})}}{
\infer[\beta]{\seqs{((\np\bs s)/\np^f\stimes\np^i)\ssl(n^{\covarc}\sbs n^m)}{\fusionpair{\lambda\fusionpair{\covarb}{b}(f \ \fusionpair{i}{\fusionpair{\lambda z(\covarb \ z)}{b}})}{\fusionpair{\lambda y(\covarc \ y)}{m}}}{(n\bs n)/(\np\bs s)}}{
\infer[R^{\circ}]{\seqs{(\np\bs s)/\np^f\stimes\np^i}{\lambda\fusionpair{\covarb}{b}(f \ \fusionpair{i}{\fusionpair{\lambda z(\covarb \ z)}{b}})}{\np\bs s}}{
\infer[\alpha]{\seq{(\np\bs s)/\np^f\stimes\np^i}{\np\bs s^{\delta}}{\casefusion{\delta}{\covarb}{b}{(f \ \fusionpair{i}{\fusionpair{\lambda z(\covarb \ z)}{b}})}}}{
\infer=[dp]{\seq{(\np\bs s)/\np^f\stimes\np^i}{s^{\covarb}\sbs\np^b}{(f \ \fusionpair{i}{\fusionpair{\lambda z(\covarb \ z)}{b}})}}{
\infer[tv]{\seq{\np^b\stimes((\np\bs s)/\np^f\stimes\np^i)}{s^{\covarb}}{(f \ \fusionpair{i}{\fusionpair{\lambda z(\covarb \ z)}{b}})}}{}}}} &
\infer[\beta]{\seqs{n^{\covarc}\sbs n^m}{\fusionpair{\lambda y(\covarc \ y)}{m}:n\bs n}}{
\infer[R^{\bullet}]{\seqs{n^{\covarc}}{\lambda y(\covarc \ y)}{n}}{
\infer[D^{\circ}]{\seq{n^{\covarc}}{n^y}{(\covarc \ y)}}{
\infer[Ax]{\seqs{n^y}{y}{n}}{}}} &
\infer[Ax]{\seqs{n^m}{m}{n}}{}}}}}$ \\ \\ \\ \\

$\infer=[dp]{\seq{n^l\stimes((n\bs n)/(s\lgplus\negl{\np})^t\stimes(\np^b\stimes(\np\bs s)/\np^r))}{n^{\covarc}}{(t \ \fusionpair{\lambda\fusionpair{\covara}{\covarb}(r \ \fusionpair{\covara}{\fusionpair{\lambda z(\covarb \ z)}{b}})}{\fusionpair{\lambda y(\covarc \ y)}{l}})}}{
\infer[D^{\bullet}]{\seq{(\np^b\stimes(\np\bs s)/\np^r)\ssl(n^{\covarc}\sbs n^l)}{(n\bs n)/(s\lgplus\negl{\np})^t}{(t \ \fusionpair{\lambda\fusionpair{\covara}{\covarb}(r \ \fusionpair{\covara}{\fusionpair{\lambda z(\covarb \ z)}{b}})}{\fusionpair{\lambda y(\covarc \ y)}{l}})}}{
\infer[\beta]{\seqs{(\np^b\stimes(\np\bs s)/\np^r)\ssl(n^{\covarc}\sbs n^l)}{\fusionpair{\lambda\fusionpair{\covara}{\covarb}(r \ \fusionpair{\covara}{\fusionpair{\lambda z(\covarb \ z)}{b}})}{\fusionpair{\lambda y(\covarc \ y)}{l}}}{(n\bs n)/(s\lgplus\negl{\np})}}{
\infer[R^{\circ}]{\seqs{\np^b\stimes(\np\bs s)/\np^r}{\lambda\fusionpair{\covara}{\covarb}(r \ \fusionpair{\covara}{\fusionpair{\lambda z(\covarb \ z)}{b}})}{s\lgplus\negl{\np}}}{
\infer[\alpha]{\seq{\np^b\stimes(\np\bs s)/\np^r}{s\lgplus\negl{\np}^{\delta}}{\casefusion{\delta}{\covara}{\covarb}{(r \ \fusionpair{\covara}{\fusionpair{\lambda z(\covarb \ z)}{b}})}}}{
\infer[(\os,\bs),dp]{\seq{\np^b\stimes(\np\bs s)/\np^r}{\negl{\np}^{\covara}\splus s^{\covarb}}{\casefusion{\delta}{\covara}{\covarb}{(r \ \fusionpair{\covara}{\fusionpair{\lambda z(\covarb \ z)}{b}})}}}{
\infer[\alpha]{\seq{(s^{\covarb}\sbs\np^b)\sobs(\np\bs s)/\np^r}{\negl{\np}^{\covara}}{(r \ \fusionpair{\covara}{\fusionpair{\lambda z(\covarb \ z)}{b}})}}{
\infer[(\obs,\negl{\cdot}),dp]{\seq{(s^{\covarb}\sbs\np^b)\sobs(\np\bs s)/\np^r}{\slgc{(\np^e)}}{(r \ \fusionpair{e}{\fusionpair{\lambda z(\covarb \ z)}{b}})}}{
\infer[tv]{\seq{\np^b\stimes((\np\bs s)/\np^r\stimes\np^e)}{s^{\covarb}}{(r \ \fusionpair{e}{\fusionpair{\lambda z(\covarb \ z)}{b}})}}{}}}}}} &
\infer[\beta]{\seqs{n^{\covarc}\sbs n^l}{\fusionpair{\lambda y(\covarc \ y)}{l}:n\bs n}}{
\infer[R^{\bullet}]{\seqs{n^{\covarc}}{\lambda y(\covarc \ y)}{n}}{
\infer[D^{\circ}]{\seq{n^{\covarc}}{n^y}{(\covarc \ y)}}{
\infer[Ax]{\seqs{n^y}{y}{n}}{}}} &
\infer[Ax]{\seqs{n^l}{l}{n}}{}}}}}$
\end{tabular}
\end{center}
\caption{Derivations of complex nouns demonstrating (peripheral) subject and non-subject extraction respectively. Words (or rather, the formulas representing their categories) appear as hypotheses, grouping together into binary branching tree structures via the structural counterpart $\stimes$ of $\lgtimes$. The chosen variable names are meant to be suggestive of the words they represent. Applications of $(tv)$ refer to Figure \ref{deriv_sample}.}
\label{deriv_relative}
\end{sidewaysfigure}

\section{Comparison}
Bernardi and Moortgat (B$\&$M, \cite{bernardimoortgat}) alternatively propose designing a Montagovian semantics for \textbf{LG} on the assumption that all formulae are of equal polarity: either all negative, inducing a call-by-name translation (CBN), or all positive, corresponding to call-by-value (CBV). Thus, the corresponding maps $\floor{\cdot}$ and $\ceil{\cdot}$ restrict to the top- and bottom levels respectively of the polarity table in $\S$2, inserting additional negations for positives in CBN and negatives in CBV:
\begin{center}
\begin{tabular}{|l|l|l|} \hline
 & $\lfloor\cdot\rfloor$ (CBN) & $\lceil\cdot\rceil$ (CBV) \\ \hline
$p$ & $\lnot p$ & $p$ \\
$A/B,B\bs A$ & $\lnot\lfloor B\rfloor\lgtimes\lfloor A\rfloor$ & $\lnot(\lceil B\rceil\lgtimes\lnot\lceil A\rceil)$ \\
$B\obs A,A\os B$ & $\lnot(\lfloor B\rfloor\lgtimes\lnot\lfloor A\rfloor)$ & $\lnot\lceil B\rceil\lgtimes\lceil A\rceil$ \\ \hline
\end{tabular}
\end{center}
B$\&$M code their derivations inside a variation of Curien and Herbelin's $\bar{\lambda}\mu\widetilde{\mu}$-calculus. However, to facilitate comparison with our own approach, we express in Figure \ref{lgt_lgq} B$\&$M's CBN and CBV translations by 'stouped' display calculi \textbf{LGT} and \textbf{LGQ}, named after their obvious sources of inspiration \cite{lkt_lkq}. Sequents, as well as their display equivalences, carry over straightforwardly from $\S$2, their interpretations being as before. In particular, atomic (co)structures are interpreted
\begin{center}
\begin{tabular}{rclcrcl}
$\floor{A^{\vara}}$ & \ $=$ \ & $\{\lnot\floor{A}^{\vara}\}$ & \ \ \ \ \ & $\floor{A^{\covara}}$ & \ $=$ \ & $\{\floor{A}^{\covara}\}$ \\
$\ceil{A^{\vara}}$ & \ $=$ \ & $\{\ceil{A}^{\vara}\}$ & & $\ceil{A^{\covara}}$ & \ $=$ \ & $\{\lnot\ceil{A}^{\covara}\}$
\end{tabular}
\end{center}
The differences between the various display calculi of Figures \ref{lgfoc_rules} and \ref{lgt_lgq} are now reduced to the maintenance of the stoup. In particular, \textbf{LGT}, considering all formulas negative, allows only hypotheses inside, whereas \textbf{LGQ} restricts the contents of the stoup to conclusions.

In comparing the various proposals at the level of the lexical semantics, the polarized approach often amounts to the more economic one. For instance, a ditransitive verb like \textit{offered}, categorized $((\np\bs s)/\np)/\np$ (abbreviated $dtv$), receives denotations of types $\lnot\floor{dtv}$ (CBN), $\ceil{dtv}$ (CBV) and $\lnot\sem{dtv}$ (polarized):
\begin{center}
\begin{tabular}{rcl}
CBN: & \ \ & $\lambda\langle Z,\langle Y,\langle X,q\rangle\rangle\rangle(Z \ \lambda z(Y \ \lambda y(X \ \lambda x(q \ (((\textsc{offered} \ z) \ y) \ x)))))$ \\
CBV: & \ \ & $\lambda\langle z,Y\rangle(Y \ \lambda\langle y,X\rangle(X \ \lambda\langle x,q\rangle(q \ (((\textsc{offered} \ z) \ y) \ x))))$ \\
polarized: & \ \ & $\lambda\langle z,\langle y,\langle q,x\rangle\rangle\rangle(q \ (((\textsc{offered} \ z) \ y) \ x))$
\end{tabular}
\end{center}
\subsubsection{Acknowledgements} This work has benefited from discussions with Michael Moortgat, Jeroen Bransen and Vincent van Oostrom, as well as from comments from two anonymous referees. All remaining errors are my own.

\begin{figure}[h]
\begin{center}
\begin{tabular}{c}
\textbf{LGT.} (Call-by-name) \\ \\
\begin{tabular}{ccc}
$\infer[Ax]{\seqs{p^{\covara}}{\covara}{p}}{}$ & \ \ \ \ \ & 
$\infer[D]{\seq{\Pi}{A^{\vara}}{(\vara \ M)}}{\seqs{\Pi}{M}{A}}$ \\ \\
$\infer[\lgtimes^{\bullet}]{\seqs{\Pi}{\lambda\fusionpair{\vara}{\varb}M}{A\lgtimes B}}{\seq{\Pi}{A^{\vara}\stimes B^{\varb}}{M}}$ & &
$\infer[\lgtimes^{\circ}]{\seq{\Gamma\stimes\Delta}{A\lgtimes B^{\covarc}}{(\covarc \ \fusionpair{\lambda\covara M}{\lambda\covarb N})}}{\seq{\Gamma}{A^{\covara}}{M} & \seq{\Delta}{B^{\covarb}}{N}}$ \\ \\
$\infer[\bs^{\circ}]{\seq{\Gamma}{B\bs A^{\covarc}}{\casefusion{\covarc}{\covara}{\varb}{M}}}{\seq{\Gamma}{A^{\covara}\ssl B^{\varb}}{M}}$ & &
$\infer[\bs^{\bullet}]{\seqs{\Pi\ssl\Delta}{\fusionpair{\lambda\covara N}{M}}{B\bs A}}{
\seq{\Delta}{B^{\covara}}{N} & \seqs{\Pi}{M}{A}}$ \\ \\
$\infer[\lgplus^{\circ}]{\seq{\Gamma}{A\lgplus B^{\covarc}}{\casefusion{\gamma}{\beta}{\alpha}{M}}}{\seq{\Gamma}{B^{\covarb}\splus A^{\covara}}{M}}$ & &
$\infer[\lgplus^{\bullet}]{\seqs{\Sigma\splus\Pi}{\fusionpair{N}{M}}{A\lgplus B}}{\seqs{\Sigma}{N}{B} & \seqs{\Pi}{M}{A}}$ \\ \\
$\infer[\os^{\bullet}]{\seqs{\Pi}{\lambda\fusionpair{\varb}{\covarc}{M}}{A\os B}}{\seq{\Pi}{A^{\varb}\sos B^{\covarc}}{M}}$ & &
$\infer[\os^{\circ}]{\seq{\Gamma\sos\Sigma}{A\os B^{\covarc}}{(\covarc \ \fusionpair{N}{\lambda\covara M})}}{\seqs{\Sigma}{N}{B} & \seq{\Gamma}{A^{\covara}}{M}}$
\end{tabular} \\ \\ \\
\textbf{LGQ.} (Call-by-value) \\ \\
\begin{tabular}{ccc}
$\infer[Ax]{\seqs{p^{\vara}}{\vara}{p}}{}$ & \ \ \ \ \ & 
$\infer[D]{\seq{\Gamma}{A^{\covara}}{(\covara \ M)}}{\seqs{\Gamma}{M}{A}}$ \\ \\
$\infer[\lgtimes^{\bullet}]{\seq{\Pi}{A\lgtimes B^{\vara}}{\casefusion{\vara}{\varb}{\varc}{M}}}{\seq{\Pi}{A^{\varb}\stimes B^{\varc}}{M}}$ & &
$\infer[\lgtimes^{\circ}]{\seqs{\Gamma\stimes\Delta}{\fusionpair{M}{N}}{A\lgtimes B}}{\seqs{\Gamma}{M}{A} & \seqs{\Delta}{N}{B}}$ \\ \\
$\infer[\bs^{\circ}]{\seqs{\Gamma}{\lambda\fusionpair{\covara}{\varb} M}{B\bs A}}{\seq{\Gamma}{A^{\covara}\sbs B^{\varb}}{M}}$ & &
$\infer[\bs^{\bullet}]{\seq{\Pi\sbs\Delta}{B\bs A^{\varc}}{(\varc \ \fusionpair{N}{\lambda\vara M})}}{\seqs{\Delta}{N}{B} & \seq{\Pi}{A^{\vara}}{M}}$ \\ \\
$\infer[\lgplus^{\circ}]{\seqs{\Gamma}{\lambda\fusionpair{\covarb}{\covara} M}{A\lgplus B}}{\seq{\Gamma}{B^{\covarb}\splus A^{\covara}}{M}}$ & &
$\infer[\lgplus^{\bullet}]{\seq{\Sigma\splus\Pi}{A\lgplus B^{\varc}}{(z \ \fusionpair{\lambda\varb N}{\lambda\vara M})}}{\seq{\Sigma}{B^{\varb}}{N} & \seq{\Pi}{A^{\vara}}{M}}$ \\ \\
$\infer[\os^{\bullet}]{\seq{\Pi}{A\os B^{\varc}}{\casefusion{\varc}{\vara}{\covarb}{M}}}{\seq{\Pi}{A^{\vara}\sos B^{\covarb}}{M}}$ & &
$\infer[\os^{\circ}]{\seqs{\Gamma\sobs\Sigma}{\fusionpair{\lambda\varc N}{M}}{A\os B}}{\seq{\Sigma}{B^{\varc}}{N} & \seqs{\Gamma}{M}{A}}$
\end{tabular}
\end{tabular}
\end{center}
\caption{Explicating the CBN and CBV interpretations of (\cite{bernardimoortgat}) through the display calculi \textbf{LGT} and \textbf{LGQ}. For reasons of space, we discuss only the binary connectives and have refrained from mentioning the display postulates (see Figure \ref{lgfoc_rules}). In addition, only rules for $\bs,\os$ are explicated, those for $/,\obs$ being similar.}
\label{lgt_lgq}
\end{figure}
\clearpage

\bibliography{lg_cps}
\end{document}